\documentclass[letterpaper, 10 pt, conference]{ieeeconf}  % Comment this line out if you need a4paper

\IEEEoverridecommandlockouts                              % This command is only needed if 
\usepackage[utf8]{inputenc}
\usepackage{xspace}
\usepackage{comment}
\usepackage{graphicx}
\usepackage{xcolor}
\definecolor{gg}{RGB}{0, 155, 85}
\definecolor{primarycolor}{RGB}{33,49,77}   % dark blue
\definecolor{angrycolor}{RGB}{210,73,42}    % orange

\usepackage{todonotes}

\usepackage{amsmath}
\usepackage{amssymb}
\usepackage{amsthm}
\theoremstyle{definition}

\makeatletter
\let\NAT@parse\undefined
\makeatother
\usepackage[pdfa,colorlinks,bookmarksopen,bookmarksnumbered,allcolors=gg]{hyperref}
\usepackage{cite}
\usepackage[algoruled,vlined,linesnumbered]{algorithm2e}

\SetAlgoSkip{SkipBeforeAndAfter}

\SetCommentSty{mycommfont}
\SetKw{Continue}{continue}
\usepackage{booktabs}
\usepackage{multirow}
\usepackage{makecell}

\usepackage[font=footnotesize]{caption}
\usepackage[font=footnotesize]{subcaption}
\usepackage[export]{adjustbox}

\usepackage[
activate   = {true},
protrusion = true,
expansion  = true,
kerning    = true,
spacing    = true,
tracking   = false,
auto       = true,
selected   = true,
factor     = 1000,
stretch    = 15,
shrink     = 15,
]{microtype}

\newcommand{\Of}{\ensuremath{O_\mathrm{f}}\xspace}
\newcommand{\Om}{\ensuremath{O_\mathrm{m}}\xspace}

\newcommand{\start}{\text{start}}

\newcommand{\goal}{\text{goal}}
\newcommand{\free}{\text{free}}

\newcommand{\C}{{\ensuremath{\mathcal{C}}}\xspace}

\newcommand{\Cg}{{\ensuremath{\C_\goal}}\xspace}
\newcommand{\qs}{{\ensuremath{q_\start}}\xspace}
\newcommand{\Cs}{\ensuremath{\mathcal{C}}-space\xspace}
\newcommand{\Cf}{{\ensuremath{\C_\free}}\xspace}
\newcommand{\Co}{{\ensuremath{\C_\text{obs}}}\xspace}

\newcommand{\graph}{\ensuremath{\mathcal{G}}\xspace}

\newcommand{\M}{\ensuremath{\mathcal{M}}\xspace}
\newcommand{\Mi}{\ensuremath{\M_i}\xspace}

\newcommand{\pv}{\ensuremath{\pi}\xspace} %a path
\newcommand{\abbr}[1]{\textsc{\MakeLowercase{#1}}\xspace}

\newcommand{\method}{\abbr{Cover}}
\newcommand{\app}{\abbr{App}}
\newcommand{\appcover}{\abbr{App-Cover}}
\newcommand{\pfs}{\abbr{PFS}}
\newcommand{\rrtconnect}{RRTConnect\xspace}

\title{\LARGE \bf
COVER: COverage-VErified Roadmaps for Fixed-time Motion Planning in Continuous Semi-Static Environments}

\author{Niranjan Kumar Ilampooranan, Constantinos Chamzas
  \thanks{%
  All authors are affiliated with the Department of Robotics Engineering, Worcester Polytechnic Institute (WPI), Worcester, MA 01609, USA {\tt\small \{nilampooranan, cchamzas\} @ wpi.edu}.
  }
}
\begin{document}

\maketitle

\thispagestyle{empty}
\pagestyle{empty}

\begin{abstract}
The ability to solve motion-planning queries within a fixed time budget is critical for deploying robotic systems in time-sensitive applications. Semi-static environments, where most of the workspace remains fixed while a subset of obstacles varies between tasks, exhibit structured variability that can be exploited to provide stronger guarantees than general-purpose planners. However, existing approaches either lack formal coverage guarantees or rely on discretizations of obstacle configurations that restrict applicability to realistic domains. This paper introduces \method, a framework that incrementally constructs coverage-verified roadmaps for semi-static environments. \method decomposes the arrangement space by independently partitioning the configuration space of each movable obstacle and verifies roadmap feasibility within each partition, enabling fixed-time query resolution for verified regions. We evaluate \method on a 7-DoF manipulator performing object-picking in tabletop and shelf environments, demonstrating broader problem-space coverage and higher query success rates than prior work, particularly with obstacles of different sizes.
\end{abstract}

\section{Introduction}

Motion planning \cite{choset2005principles} is a core function of robotics, enabling manipulators to plan collision-free, reliable trajectories in complex workspaces. In industrial domains, robots often operate in semi-static environments—settings where the majority of the workspace remains fixed, while smaller obstacles such as bins and packages vary across tasks. These environments are characterized by the repetitive nature of the tasks performed, such as moving similar objects between shelves or tables. In such scenarios, robots must generate motions within strict time budgets while minimizing downtime.  

\begin{figure}
    \centering
    \includegraphics[width=0.95\linewidth]{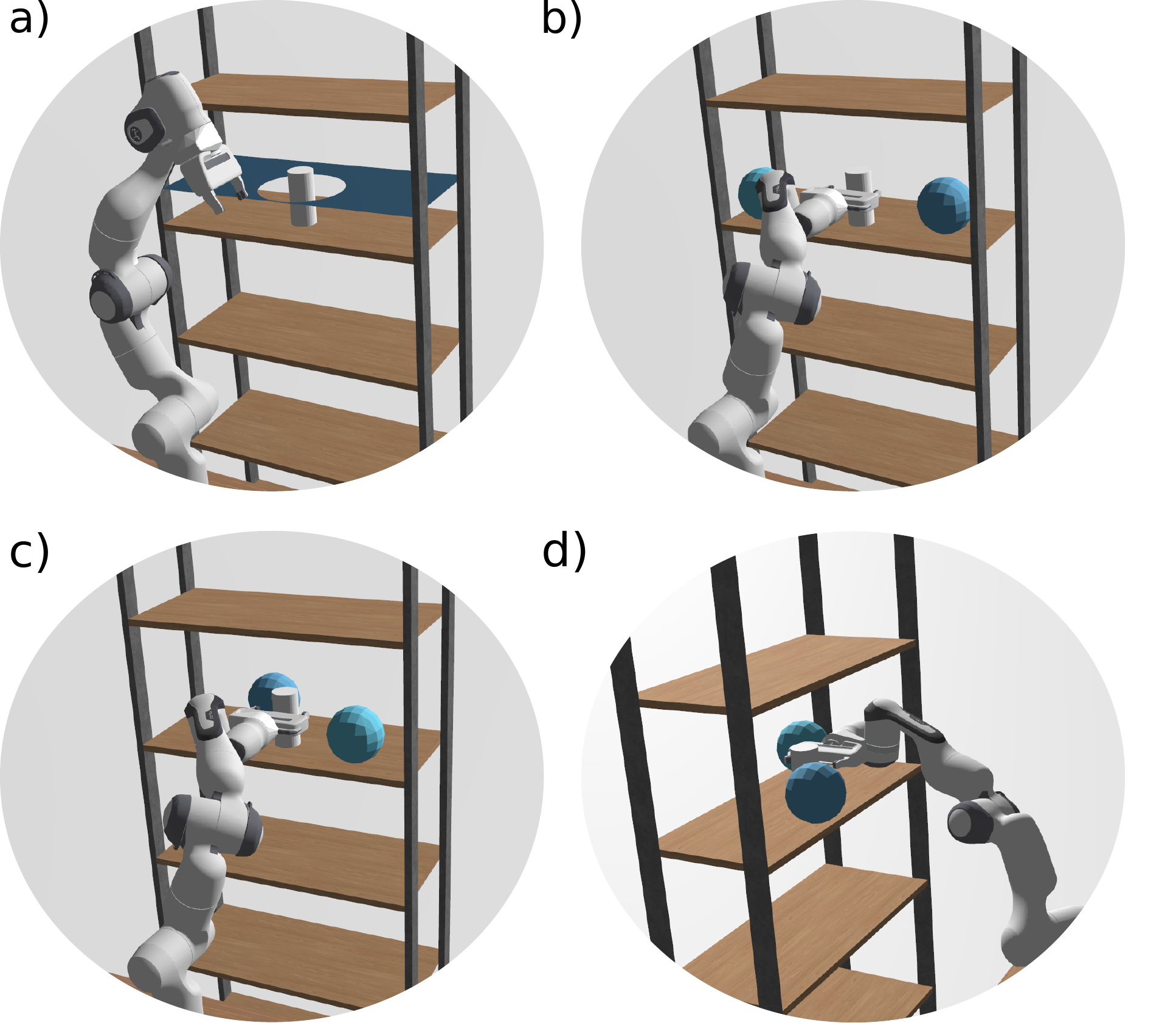}
    \caption{\textbf{Semi-static motion-planning problem.} \textbf{(a)} The robot must grasp a white cylinder while two spherical obstacles may lie anywhere in the dark-blue region. \textbf{(b)–(d)} Three sampled instances for the movable obstacles. \method finds a valid motion plan for the covered regions in fixed time.}
    \vspace{-1.5 em}
    \label{fig:semi-static}
\end{figure}

For instance, consider the scenario illustrated in \autoref{fig:semi-static}, where a robot must retrieve a cylindrical object from a shelf while two additional movable obstacles can occupy arbitrary positions within the same shelf region. Such semi-static variability frequently arises in real-world settings, particularly in warehouse and industrial environments, where bins and packages are repeatedly rearranged between tasks.

Preprocessing-based methods exploit this structured variability to accelerate query-time planning and, in some cases, achieve fixed-time performance. Experience-based frameworks \cite{coleman_experience-based_2014, berenson_robot_2012, chamzas_learning_2021, ishida_coverlib_2025, phillips_e-graphs_2021} improve efficiency by reusing prior solutions but provide no coverage guarantees—that is, they cannot certify whether a valid obstacle arrangement will admit a solution. Fixed-time motion planners \cite{islam_alternative_2021, van_den_berg_creating_2005} do provide formal guarantees, but require restrictive assumptions such as discretizing the obstacle configuration space and assuming all movable obstacles to be identical. These assumptions limit their applicability to realistic domains, where obstacles vary continuously in both size and placement.

%Brief description of the method
We propose \method, a roadmap-coverage framework for fixed-time motion planning that extends beyond discretized formulations to handle continuous problem spaces while maintaining fixed-time guarantees. Given a semi-static environment represented as a continuous set of possible obstacle placements, \method incrementally constructs a roadmap in the robot’s configuration space. By associating robot motions on the roadmap with obstacle configurations that invalidate them, \method systematically verifies coverage of the problem space, enabling fixed-time query resolution even when obstacle configurations are defined over continuous domains.

We evaluate \method using a 7-DoF manipulator performing object-picking tasks in tabletop and shelf environments. Our experiments show that \method achieves significantly broader problem-space coverage than state-of-the-art fixed-time baselines, particularly in heterogeneous obstacle setups. Furthermore, \method resolves complex motion queries orders of magnitude faster than single-query planning from scratch, while maintaining high success rates even in narrow-passage scenarios where traditional planners often fail to find a solution within reasonable time limits.

Our main contributions are as follows: (i) we introduce \method which provides fixed query-time guarantees in continuous obstacle spaces, (ii) demonstrate its ability to handle obstacles of varying sizes and continuous placements, and (iii) develop a problem-space coverage estimator that verifies the exact coverage ratio of precomputed paths or roadmaps.

\section{Related Work}

There have been several motion planning approaches that leverage preprocessing to accelerate query-time performance. The most well-known are roadmap planners \cite{kavraki1996probabilistic, bohlin_path_2000, marcucci_motion_2023}, which construct a roadmap of collision-free configurations during an offline phase and then answer multiple queries on the same graph. While highly effective in static environments, these methods degrade when obstacles change, as the roadmap may no longer reflect the true free space and must be repaired.

To address this limitation, the class of dynamic roadmaps was proposed. Such methods extend multi-query planning to environments that change between queries by maintaining mappings from workspace regions to configuration-space edges \cite{kallman_motion_2004}. When the environment changes, these mappings allow for rapid invalidation of affected edges, enabling efficient online updates. This has been implemented using hashing techniques for fast collision detection \cite{leven_framework_2002, van_den_berg_roadmap-based_2005} or hierarchical decompositions for scalable updates \cite{yang_hdrm_2018}. Although highly efficient, these methods are designed for arbitrary dynamic environments—any workspace subregion can potentially be occupied — and therefore do not target the structured variability of semi-static settings. As a result, they emphasize fast updates rather than certifying coverage. 

In the class of learning-based methods, experience-based planners accelerate planning by reusing information from previously solved problems. This information may take the form of a library of paths \cite{berenson_robot_2012, ishida_coverlib_2025}, a roadmap enriched with prior experience \cite{coleman_experience-based_2014, phillips_e-graphs_2021}, or a collection of local samplers that bias future queries \cite{chamzas_learning_2021}. These approaches generally assume that planning problems are drawn from an unknown distribution and attempt to maintain a representative set of solutions or sampling strategies that can be adapted to new queries \cite{pairet_path_2021, ishida_coverlib_2025}. While effective in practice, they do not explicitly exploit the structured variability of semi-static environments. Instead, they focus on statistical generalization across distributions of problems, without offering guarantees of full coverage in the space of obstacle configurations.

Fixed-time motion planning methods \cite{van_den_berg_creating_2005, islam_alternative_2021, mishani2024constant} are most closely related to our work. These approaches specifically target semi-static environments by assuming known apriori distributions of movable obstacles and guarantee that queries can be answered within a fixed-time budget through extensive offline preprocessing. However, existing formulations discretize the obstacle arrangement space, which restricts their applicability in realistic scenarios where obstacles can vary continuously. Moreover, recent formulations \cite{islam_alternative_2021} impose strong homogeneity assumptions, such as requiring all obstacles to occupy equivalent locations or share identical shapes, further limiting their use in heterogeneous settings.

\section{Problem Statement and Notation}

\subsection{Geometric Motion Planning} 
Let $q \in \mathcal{C}$ denote a configuration and configuration space (\Cs) of the robot. The set of robot configurations in collision with obstacles in the workspace is given by: 
\[
\Co = \{q \in \mathcal{C} \mid R(q) \cap \mathcal{WO} \neq \emptyset \}
\]
where $R(q) \subseteq \mathbb{R}^3$ is the workspace occupancy of the robot at configuration $q$ and $\mathcal{WO} \subseteq \mathbb{R}^3$ denotes the occupancy of obstacles in the workspace.\footnote{Self-collisions or kinematic constraints can be encoded similarly.}
Conversely, valid robot configurations in \Cs are defined as $\Cf = \C \setminus \Co$.

Given $\qs \in \Cf$ as the start configuration, $\Cg \subseteq \Cf$  as the goal region, and $\mathcal{WO}$ as the workspace occupancy of the obstacles, a geometric motion-planning  problem instance is specified by $p=(\qs,\Cg,\Cf)$. 
The task is to find a continuous path $\pv:[0,1]\to\Cf$ such that $\pv(0)=\qs$ and $\pv(1)\in\Cg$, if one exists.

% \subsection{Geometric Motion Planning in Semi-Static Workspaces}   
% A semi-static workspace contains two disjoint sets of obstacles: fixed (static) obstacles $\Of$ and movable obstacles $\Om=\{o_1,o_2,\ldots,o_n\}$ whose configurations may vary across planning queries but remain fixed during execution. Each arrangement $m$ of the movable obstacles from the space of possible arrangements $\mathcal{M}$ induces a distinct collision-free space, $\Cf(m)$ and thus a distinct geometric motion-planning instance given in \autoref{eqn:problem-semi-static}.
% \begin{equation}
%     p(m) = (\qs,\Cg, \Cf(m))
%     \label{eqn:problem-semi-static}
% \end{equation}
% Here, $\mathcal{WO}(m)$ indicates the total occupancy of the obstacles in the workspace, including those of movable obstacles after assuming an arrangement $m \in \mathcal{M}$, denoted as $O_m(m)$. Now, given a start configuration $q_s$, and a goal region, $\Cg$ the \emph{semi-static motion-planning problem} is to find, for every $m \in \M$, a continuous path $\pv_m:[0,1]\to \Cf(m)$ such that $\pv_m(0)=\qs$ and $\pv_m(1)\in\Cg$, if one exists.

\subsection{Geometric Motion Planning in Semi-Static Workspaces}   
A semi-static workspace contains two disjoint sets of obstacles: fixed (static) obstacles \Of ~and movable obstacles $\Om = \{o_1, o_2, \dots, o_n\}$. We denote the joint configuration space of these movable obstacles as the \emph{arrangement space} $\mathcal{M}$, where an arrangement $m \in \mathcal{M}$ specifies the configuration of every obstacle in $\mathcal{O}_m$. 

In such semi-static environments, the total workspace occupancy $\mathcal{WO}(m) = \mathcal{WO}_f \cup \mathcal{WO}_m(m)$ is determined by the specific arrangement $m$, where $\mathcal{WO}_f$ and $\mathcal{WO}_m(m)$ denote the workspace occupancies of $\Of$ and $\Om$ assuming arrangement $m$, respectively. This occupancy induces a distinct configuration-space obstacle region:
\[
    \Co(m) = \{q \in \C \mid R(q) \cap \mathcal{WO}(m) \neq \emptyset\}
\]

As a result, the arrangement space $\M$ parametrizes a family of motion-planning instances $p(m) = (\qs, \Cg, \Cf(m))$, where $\Cf(m) = \C  \setminus  \Co(m)$. Now, given a start configuration $\qs$ and a goal region $\Cg$, the \emph{semi-static motion-planning problem} is to find for each $m \in \M$, a continuous path $\pv_m:[0,1]\to \Cf(m)$ such that $\pv_m(0)=\qs$ and $\pv_m(1)\in\Cg$, if one exists.

\section{Methodology}

Exhaustively solving the motion-planning problem for every arrangement $m \in \M$ in a semi-static workspace is infeasible, since the space of movable obstacle configurations is continuous and therefore infinite. Instead, \method builds on the key observation that a single solution path can remain valid across infinitely many obstacle arrangements. For instance, in \autoref{fig:semi-static}(b) and \autoref{fig:semi-static}(c), the same precomputed path remains feasible despite variations in obstacle placement. Leveraging this observation, \method introduces a framework for formally quantifying the portion of $\M$ a given roadmap can solve, which we refer to as \emph{problem-space coverage}.

% Section~\ref{sec:ps-coverage} formalizes this concept and outlines the key ideas behind the computation for roadmaps in semi-static environments. Building on this foundation, Section~\ref{sec:build} presents our coverage-informed roadmap construction algorithm, which incrementally maximizes problem-space coverage by systematically addressing uncovered regions of the obstacle configuration space. Finally, Section~\ref{sec:query} describes how the resulting \method roadmap enables fixed-time query resolution for arbitrary arrangements of movable obstacles.

% \autoref{sec:ps-coverage} formalizes the problem-space coverage metric, while \autoref{sec:build} introduces the iterative procedure for constructing coverage-informed roadmaps by resolving composite motion-planning problems. \autoref{sec:decomposition} details the use of Binary Space Partitioning (BSP) trees to decompose obstacle configuration spaces into regions of equivalent edge-obstruction signatures. Finally, \autoref{sec:query} and \autoref{sec:theoretical-guarantees} describe the query-time path retrieval process and the theoretical guarantees afforded by the \method framework.

\subsection{Problem Space Coverage Estimation}\label{sec:ps-coverage}
Formally, a \emph{roadmap} is defined as a graph $\graph = (V, E)$ in the robot's configuration space $\Cf$~\cite{orthey_sampling-based_2024}. Each vertex $v \in V$ corresponds to a robot configuration $q \in \Cf$, while each edge $e \in E$ represents a collision-free motion between two vertices. Solving a query between two configurations amounts to searching for a sequence of edges in $\graph$ that connects the corresponding vertices. Roadmaps provide a compact representation of feasible motions and form the basis of multi-query motion planning~\cite{kavraki1996probabilistic, dobson_sparse_2014, coleman_experience-based_2014}.

As \Cf remains unchanged in static environments, roadmap vertices and edges retain their validity, and therefore any previously feasible path also remains valid. However, in semi-static environments, the collision-free space $\Cf(m)$ depends on the specific arrangement $m$ of movable obstacles $\mathcal{O}_m$. For a given roadmap $\mathcal{G}$, we define $\mathcal{M}_{\text{cov}} \subseteq \mathcal{M}$ in the arrangement space that admits a collision-free path from start to goal. Its complement, $\mathcal{M}_{\text{uncov}} = \mathcal{M} \setminus \mathcal{M}_{\text{cov}}$, consists of arrangements that invalidate all collision-free paths in \graph. Thus, the problem-space coverage of \graph (\autoref{eqn:problem_space}) is defined as:
% Thus, the \emph{problem-space coverage} of 
% \graph (\autoref{eqn:problem_space}) is defined as the proportion of obstacle arrangements in $\M$, that do not invalidate all paths, given by:
\begin{equation}
    \text{Coverage}(\graph) 
    = \frac{\text{Vol}(\mathcal{M}_{\text{cov}})}{\text{Vol}(\mathcal{M})},
    \label{eqn:problem_space}
\end{equation}

where $\text{Vol}(\cdot)$ denotes the measure of a subregion. This metric captures the proportion of the semi-static problem space guaranteed to be solvable with \graph. Note that some subregions of $\M$ may correspond to inherently infeasible problems \cite{li_sampling_2023}.

% \todo{Explain what s-t paths is or just remove s-t paths}

\subsection{Building the Roadmap}\label{sec:build}

\method constructs a coverage-informed roadmap through an iterative procedure that incrementally expands the portion of the arrangement space guaranteed to be solved (\autoref{alg:cover}). At each iteration, the algorithm identifies an uncovered subregion $s \subseteq \M$ and formulates a \emph{composite} motion-planning problem $p(s) = \{ p(m) \mid m \in s \}$. Intuitively, $p(s)$ corresponds to planning with the \emph{joint obstacle occupancy} $\mathcal{WO}(s)$ induced by arrangements in $s$, where $\mathcal{WO}(s) = \bigcup_{m \in s} \mathcal{WO}(m)$. The roadmap is then augmented with new edges and vertices to ensure a path that is collision-free for every arrangement $m \in s$. An example of this process is illustrated in \autoref{fig:methodology-fig}.

\begin{algorithm}[!t]
\caption{Computing \method}
\label{alg:cover}
\SetKwInput{KwInput}{Input}
\SetKwInput{KwOutput}{Output}

\KwInput{Starts $Q_s$, Goals $Q_g$, Static $O_f$ and Movable obstacles $O_m$, Planner $\mathcal{P}$}
\KwOutput{Roadmap \graph, Binary Space Partitioning trees $T$, Unverified obstacle arrangements $S$}

\SetKwFunction{FInit}{InitializeRoadmap}
\SetKwFunction{FPartition}{PartitionObstacleSpace}
\SetKwFunction{FRemoveInvalid}{RemoveInvalidEdges}
\SetKwFunction{FFindPath}{FindPath}
\SetKwFunction{FRepair}{RepairRoadmap}
\SetKwFunction{FAddPath}{AddPathToRoadmap}
\SetKwFunction{FAllPaths}{GetAllPaths}
\SetKwFunction{FDetect}{UncoveredArrangements}

\DontPrintSemicolon
\LinesNumbered

$\graph \gets \FInit(Q_s, Q_g)$ \label{line:init_graph}\; 

% \tcp{Generate Decomposition trees for obstacle configuration spaces}
% $T \gets \FPartitionObstacleSpace(G, O_m$) \;

\tcp{Solve a multi-start multi-goal motion-planning problem assuming only static obstacles in workspace}
% \tcp{$O_m(c)$ indicates that obstacles $O_m$ assume configurations in $c$}
$\pv \gets \FFindPath(Q_s, Q_g, O_f, \mathcal{P})$ \label{line:init_path}\; 

$\FAddPath(\graph, \pv)$ \; \label{line:add_path}

\tcp{Identify subset of arrangements currently uncovered by \graph}
$\{S, T\}\gets \FDetect(\graph, O_m)$\label{line:init_filter}\; 
\textit{failures} $\gets 0$\;

% --- Main loop
\While{$S \neq \emptyset$}{ \label{line:iter_loop_start}

  pop $s$ from $S$ \label{line:pop_s}\; 
  \tcp{Remove invalid edges due to movable obstacle arrangements in $s$}
  $\graph' \gets \FRemoveInvalid(\graph, s)$ \label{line:clone_graph}\; 

  \tcp{Solve motion-planning problem with movable obstacles assuming all arrangements in $s$}
  $\pv' \gets \FRepair(\graph', O_f, O_m(s), \mathcal{P})$\label{line:iter_repair}\; 
  \If{$\pv' \neq \emptyset$}{
     % \tcp{Adding path if planning is successful}
    $\FAddPath(\graph, \pv')$\; 
    \tcp{Remove arrangements covered by $\graph$ after adding new path}
    $\{S, T \} \gets \FDetect(\graph, O_m)$ \label{line:iter_filter}\;
    \textit{failures} $\gets 0$\;
  }
  \Else{
    add $s$ back to $S$ \;
    \textit{failures} $\gets$ \textit{failures} $+ 1$ \label{line:fail_inc}\;
    \If{\textit{failures} $== |S|$}{ \label{line:fail_threshold}
     % \tcp{Cannot resolve further arrangements} 
      \Return{$\graph, S, T$} \label{line:iter_loop_end}\;
    }
  }
}
% \tcp{Valid paths obtained for all obstacle arrangements \M}
\Return{$\graph, \emptyset, T$} \label{line:success_cond_exit}
\end{algorithm}

We begin by initializing the roadmap $\graph$ with two vertex sets consisting of robot's start and goal configurations, respectively (\autoref{line:init_graph}). Following this initialization, we attempt to find a valid path \pv to connect these vertex sets using a multi-start multi-goal planner $\mathcal{P}$\footnote{\rrtconnect is used as the underlying multi-start multi-goal motion planner in our experiments.}, assuming only the presence of static obstacles in the workspace (\autoref{line:init_path}). By adding this path to \graph, the roadmap covers the subregion $s \in \M$ for which \pv remains collision-free. In \autoref{line:init_filter}, we identify the set of uncovered subregions $S$ by decomposing each obstacle configuration space \Mi as a Binary Space Partitioning (BSP) trees $T= \{T_1, T_2, \ldots, T_n\}$, described further in \autoref{sec:decomposition}.

\begin{figure*}
    \centering
    \vspace{0.5cm}
    \includegraphics[width=\linewidth]{imgs/cover-new.pdf}
\caption{\textbf{Demonstration of \method with two movable obstacles $\mathbf{o_1}$ and $\mathbf{o_2}$}. 
\textbf{(a)} The roadmap $\graph$ is initialized with start and goal configurations, and the obstacle arrangement spaces $\mathcal{M}_1$ and $\mathcal{M}_2$ are partitioned into subregions, resulting in distinct subdivisions due to different obstacle sizes.
\textbf{(b)} A path (green edge) is computed assuming no movable obstacles and added to the roadmap, after which the partitions are updated to account for the newly added edges.
\textbf{(c)} A composite motion-planning problem is formed using joint workspace occupancy resulting from obstacle configurations that invalidate all paths in the current roadmap (e.g., path discovered in \textbf{(b)}). Nodes and edges in $\graph$ invalidated by this joint workspace occupancy are marked in red. A feasible path is then added to $\graph$, expanding its coverage. The process repeats until $\mathcal{M}_{\text{uncov}}$ is exhausted or termination threshold is reached.}
\label{fig:methodology-fig}
\vspace{-15pt}
\end{figure*}

At each iteration of the repair loop (\autoref{alg:cover}, \autoref{line:iter_loop_start}--\autoref{line:iter_loop_end}), the algorithm pops a subregion $s \in S$ (\autoref{line:pop_s}). We then extract a subgraph $\mathcal{G}' \subseteq \mathcal{G}$ consisting only of the edges and vertices that remain collision-free for all arrangements in $s$ (\autoref{line:clone_graph}). In $\mathcal{G}'$, the start and goal vertex sets belong to disjoint connected components. We attempt to restore connectivity between them by solving the composite problem $p(s)$ using the planner $\mathcal{P}$. The resulting path $\pv'$, which is valid for all arrangements in $s$, is then added to the main roadmap $\mathcal{G}$ (\autoref{line:iter_repair}). Once $\pv'$ is added, we repeat the procedure of identifying remaining uncovered arrangements by \graph after augmenting with $\pv'$ (\autoref{line:iter_filter}). This update identifies any other subregions in $S$ that are now also connected by the updated roadmap, allowing them to be removed from the set. This ensures that a single successful repair can resolve multiple subregions simultaneously.

 % Conversely, if no path can be found for the composite problem $p(s)$, the subregion $s$ remains unresolved and is returned to the set $S$. This outcome is recorded by incrementing the \textit{failures} counter before proceeding to the next subregion. This process repeats until $S$ is empty, signifying that \graph provides full coverage of the arrangement space, or until the counter reaches the termination threshold $|S|$. While further recursive subdivision of each subregion on each unsuccessful attempt could be performed to improve coverage, we restrict the scope to this consecutive termination criteria to maintain computational tractability. Upon termination, the algorithm returns the roadmap $\mathcal{G}$, the partitioning trees $T$, and any remaining subregions in $S$. 

Conversely, if no path is found for the composite problem $p(s)$, the subregion $s$ remains unresolved and is returned to $S$ while the \textit{failures} counter is incremented (\autoref{line:fail_inc}). This process repeats until $S$ is empty, signifying that \graph provides full coverage of the arrangement space, or until the counter reaches the termination threshold $|S|$ (\autoref{line:fail_threshold}). While recursive subdivision of unresolved subregions could further improve coverage, we omit this refinement to maintain computational tractability. Upon termination, the algorithm returns the roadmap $\graph$, the BSP trees $T$, and any remaining subregions in $S$.

\subsection{BSP Trees of Obstacle Configuration Spaces}
\label{sec:decomposition}

Every time a path is added to \graph in \autoref{alg:cover}, \method must update the uncovered arrangement subregions $S$. A direct approach would involve online collision checking of robot motions or edges $e \in E$, under workspace occupancy $\mathcal{WO}(s)$ for $s \in S$. However, this prohibitive operation can be avoided by precomputing a mapping from subregions $s \subseteq \M$ to the set of edges in $E$ they invalidate.

\begin{figure}
    \centering
    \includegraphics[width=\linewidth]{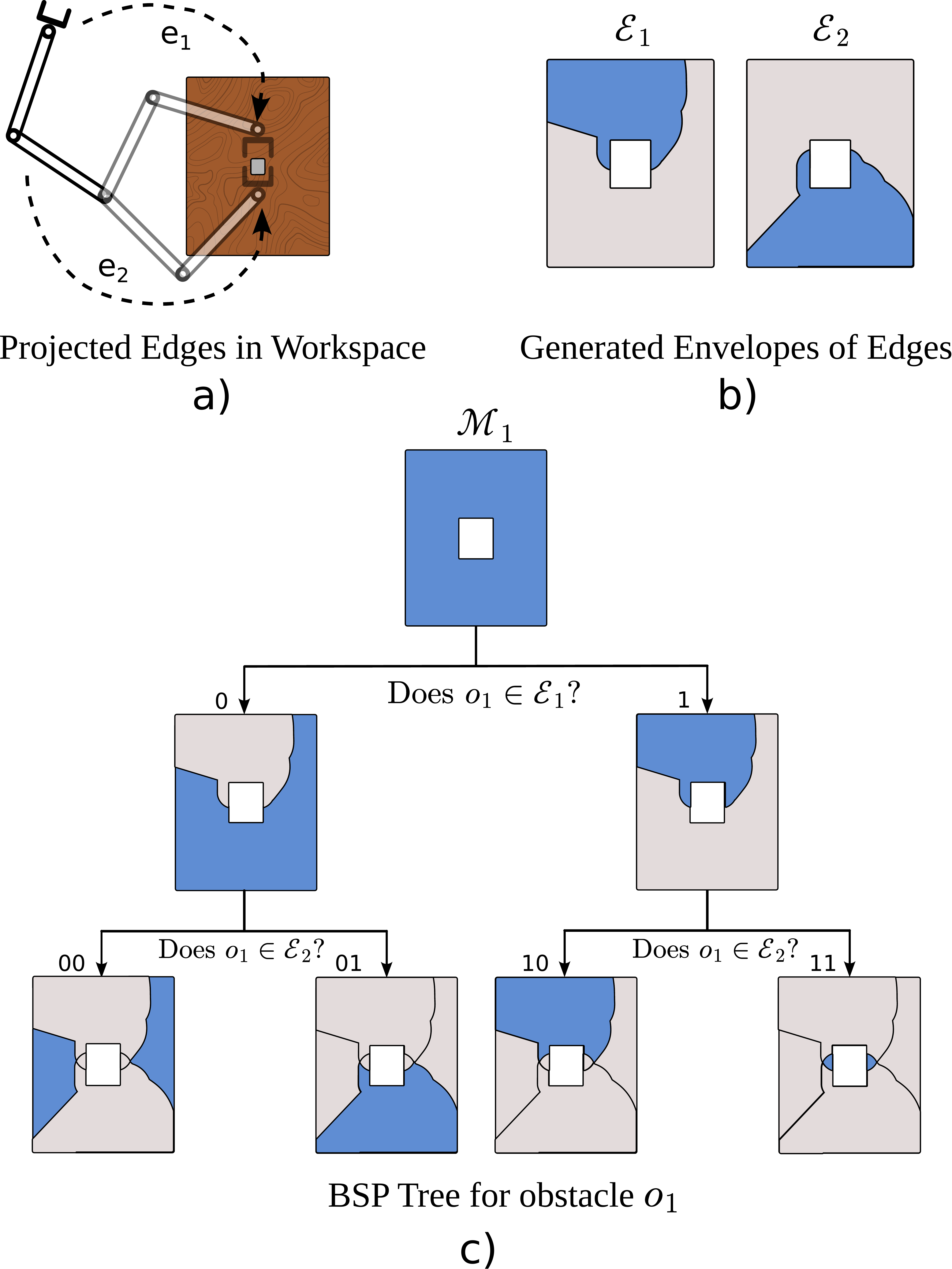}
    \caption{\textbf{(a)} Example roadmap containing two edges, where dashed curves indicate the approximate end-effector motion along each edge.
\textbf{(b)} Regions of obstacle placements that would invalidate the corresponding motions.
\textbf{(c)} These regions recursively partition the obstacle configuration space of $o_i$ into disjoint subregions organized as a binary space partitioning (BSP) tree. Each leaf corresponds to a region of placements that share the same edge-validity pattern.} 
\vspace{-2em}
    \label{fig:tree}
    \vspace{-2pt}
\end{figure}

For each obstacle $o_i$, we determine the configurations that invalidate a roadmap edge $e=(q_1,q_2)\in E$. Let $SV(e) = \bigcup_{q\in[q_1,q_2]} R(q)$ denote the robot's swept volume along $e$. The set of configurations of $o_i$ that invalidate $e$ is obtained via the Minkowski sum of $SV(e)$ with the geometry of $o_i$ as shown in \autoref{eqn:envelope}:

\begin{equation}
\mathcal{E}_e \;=\; SV(e)\;\oplus\;\mathrm{geom}(o_i),
\label{eqn:envelope}
\end{equation}

where $\oplus$ denotes the Minkowski sum and $\mathrm{geom}(o_i)$ denotes the geometry of obstacle $o_i$, known \emph{a priori}. We refer to $\mathcal{E}_e$ as the \emph{envelope} of edge $e$\footnote{The same construction can also be applied to start and goal vertices to identify obstacle configurations that invalidate them.} for obstacle $o_i$, following~\cite{islam_alternative_2021}. The envelopes $\mathcal{E}=\{\mathcal{E}_e \mid e \in E \}$ serve as splitting surfaces that recursively partition the obstacle configuration space $\mathcal{M}_i$ into disjoint subregions. Each envelope induces a binary split of $\mathcal{M}_i$ into placements that invalidate the corresponding edge and those that do not, yielding a BSP tree $T_i$. This process of recursively partitioning $\mathcal{M}_i$ using envelopes is illustrated in \autoref{fig:tree}.

The leaves of the BSP tree correspond to regions of obstacle configurations that invalidate the same subset of edges in \graph. The subregion in $\mathcal{M}_i$ represented by each leaf node therefore admits a binary encoding, referred to as a \emph{signature}, represented as a vector $\mathbf{b}_i \in \{0,1\}^{|E|}$. The $j$-th entry of $\mathbf{b}_i$ indicates whether edge $e_j \in E$ is invalidated ($1$) or remains valid ($0$) when obstacle $o_i$ is placed within that region. For instance, in the two-edge roadmap shown in \autoref{fig:tree}, the signature \texttt{11} indicates that both edges are invalidated, whereas \texttt{00} implies that neither is affected. This signature-based representation allows obstacle placements to be reasoned about algebraically rather than geometrically, enabling efficient combination of partitioned subregions across each $\M_i$ when computing $\M_\mathrm{uncov}$.

\subsection{Calculating $\mathcal{M}_{uncov}$} \label{subsec:vol}

We now use the per-obstacle signatures to identify arrangement subregions that invalidate all paths in the roadmap (\autoref{alg:validity}). For multiple obstacles, the signatures corresponding to their placements are combined into a composite signature
\[
\mathbf{b} = \mathbf{b}_1 \vee \mathbf{b}_2 \vee \cdots \vee \mathbf{b}_n,
\]
where $\vee$ denotes element-wise logical OR, indicating which edges $e_j \in E$ are invalidated by subregion $s$.

We extract the set of all paths $\Pi = \{\pi_1,\ldots,\pi_p\}$ in \graph (\autoref{alg:validity}, \autoref{line:dfs}). Each path is then encoded as a binary vector, indicating which edges belong to that path (\autoref{line:pathvector}), forming the path--edge incidence matrix $\mathbf{P} \in \{0,1\}^{p \times |E|}$.

To determine whether a composite signature $\mathbf{b}$ invalidates all paths in \graph (\autoref{line:pb}), we evaluate the matrix--vector multiplication shown in \autoref{eqn:dot-product}:

\begin{equation}
    \mathbf{v} = \mathbf{P}\mathbf{b}
    \label{eqn:dot-product}
\end{equation}

The entry $v_j$ counts the number of invalidated edges along path $\pi_j$. If $v_j > 0$ for every path $\pi_j \in \Pi$, then each path contains at least one blocked edge and the corresponding arrangement subregion is classified as uncovered. Otherwise, if there exists some $j$ such that $v_j = 0$, the roadmap still contains a valid path for that region. Iterating over all composite signatures (\autoref{line:leafloop}--\autoref{line:return}) therefore yields the uncovered subregions $S$, corresponding to $\mathcal{M}_{\mathrm{uncov}}$.

\begin{algorithm}
\caption{UncoveredArrangements}
\label{alg:validity}
\DontPrintSemicolon
\LinesNumbered
\SetKwInput{KwInput}{Input}
\SetKwInput{KwOutput}{Output}
\SetKwFunction{FDepthFirstSearch}{DepthFirstSearch}
\SetKwFunction{FPartitionObstacleSpace}{PartitionObsSpace}
\SetKwFunction{FBinaryVector}{BinaryVector}
\SetKwFunction{FCompositeSignature}{CompositeSignature}

\KwInput{Roadmap $\graph=(V,E)$, Movable obstacles $\Om$}
\KwOutput{Uncovered subregions $S$, BSP trees $T$}

\tcp{Extract all paths in the roadmap}
$\Pi \gets \FDepthFirstSearch(\graph)$\; \label{line:dfs}

\tcp{Get BSP tree for each obstacle}
$T \gets \{\FPartitionObstacleSpace(\graph,o_i)\mid o_i \in \Om\}$\; \label{line:trees}

\tcp{Build path-edge incidence matrix}
\ForEach{$\pi_j \in \Pi$}{
    $\mathbf{P}[j] \gets \FBinaryVector(\pi_j,E)$\; \label{line:pathvector}
}

$S \gets \emptyset$\; \label{line:initS}

\tcp{Each $s$ corresponds to selecting one leaf node from every BSP tree}
\ForEach{$s \in \text{Leaves}(T_1)\times\cdots\times\text{Leaves}(T_n)$}{ \label{line:leafloop}
    $\mathbf{b} \gets \FCompositeSignature(s)$\; \label{line:signature}
    $\mathbf{v} \gets \mathbf{P}\mathbf{b}$\; \label{line:pb}
    \If{$\forall j:\; v_j > 0$}{ \label{line:uncoveredcheck}
        $S \gets S \cup \{s\}$\; \label{line:addS}
    }
}

\Return{$S, T$}\; \label{line:return}
\end{algorithm}

\subsection{Querying the Roadmap}\label{sec:query}

After computing the coverage-informed roadmap $\graph$ and the BSP trees for each obstacle's configuration space, collision-free paths can be retrieved for any obstacle arrangement $m \in \M_{\text{cov}}$. Since the BSP trees already encode which roadmap edges are invalidated by obstacle placements, online collision checking against robot geometry is no longer required.

For a given obstacle arrangement $m$, querying follows the same signature evaluation procedure used in \autoref{alg:validity}, but applied to a single arrangement rather than all combinations of BSP leaf nodes. For each movable obstacle $o_i \in \Om$, the corresponding BSP tree $T_i$ is traversed to locate the leaf node containing its configuration. The binary signature $\mathbf{b}_i$ associated with that leaf specifies the subset of roadmap edges invalidated by the placement of $o_i$ (\autoref{alg:validity},~\autoref{line:signature}).

The signatures obtained from the individual trees are then combined to form the composite signature $\mathbf{b}$, which encodes the complete set of edges blocked by the obstacle arrangement $m$. This composite signature is evaluated using the matrix--vector multiplication in \autoref{eqn:dot-product} (\autoref{alg:validity}, \autoref{line:pb}). Any path $\pi_j \in \Pi$ satisfying $v_j = 0$ remains collision-free under the arrangement $m$ and can therefore be returned directly as the solution to the query.

\subsection{Theoretical Guarantees} \label{sec:theoretical-guarantees}

\textbf{Fixed Time Guarantees.} Querying the roadmap can be performed within a fixed-time budget $t_{\text{query}}$. For a given obstacle arrangement $m$, answering a query requires two operations.

First, the BSP tree $T_i$ of each movable obstacle $o_i \in \Om$ is traversed to locate the leaf node containing its configuration, yielding the corresponding binary signature $\mathbf{b}_i$. Each traversal requires $O(\log |E|)$ time. For $n$ movable obstacles, computing the composite signature $\mathbf{b}$ therefore requires $O(n \log |E|)$ time. Second, the composite signature is evaluated against the precomputed path--edge incidence matrix $\mathbf{P}$ using \autoref{eqn:dot-product}. Since $\mathbf{P} \in \{0,1\}^{|\Pi|\times |E|}$, this step requires $O(|\Pi| \cdot |E|)$ time. Thus, the time complexity of answering a query is: 

\[
O\!\left(n\log|E| + |\Pi|\cdot|E|\right)
\]

Since \graph, $T$, and $\mathbf{P}$ are constructed offline and remain fixed across obstacle arrangements, the query computation can be bounded by a predetermined time budget $t_{\text{query}}$.

\textbf{Completeness Guarantees.} These follow directly from the computed coverage of the arrangement space. If an obstacle arrangement $m \in \mathcal{M}_{\mathrm{cov}}$, then by construction there exists at least one path $\pi_j \in \Pi$ such that $v_j = 0$ in \autoref{eqn:dot-product}. Consequently, a valid collision-free path is guaranteed to exist in the roadmap and can be returned by the query procedure. If $m \notin \mathcal{M}_{\mathrm{cov}}$, then no valid paths exist in $\graph$ for $m$.
\section{Experiments}

In this section, we evaluate \method on object-picking tasks with a 7-DoF Franka Emika Panda manipulator across different simulated environments with varying task difficulty. We assess its problem-space coverage capabilities against a fixed-time baseline for semi-static environments: Alternative Paths Planner (\app)~\cite{islam_alternative_2021}. Finally, the query-time performance of \method and \app is benchmarked for multiple motion planning queries against planning-from-scratch (\pfs) using a well-tuned  \rrtconnect~\cite{kuffner_rrt-connect_2000}.

\subsection{Experimental Setup}

% All experiments were conducted in Genesis~\cite{authors_genesis_2024} simulation on a workstation with 56-core Intel Xeon CPU and 128 GB RAM. For motion planning and roadmap-based operations, we leverage the Open Motion Planning Library (OMPL) \cite{sucan_open_2012}. Furthermore, we obtain the spherical approximation of the robot links using \cite{coumar_foam_2025} and compute the corresponding swept-spheres convex hulls as an approximation of the volume swept by the robot trajectory. Since the robot occupancy and obstacle configuration spaces are modeled as polygonal meshes, we employ the methods in MeshLib \cite{noauthor_meshlib_2025} for computing their mesh representation and for the 3D Boolean operations for subdividing of obstacle configuration spaces. 

\textbf{Implementation.} All experiments are conducted in simulation using the Genesis physics platform~\cite{authors_genesis_2024}, hosted on a workstation with a 56-core Intel Xeon CPU and 128 GB RAM. OMPL \cite{sucan_open_2012} provides the underlying framework for all motion planning and roadmap-based operations. For computing swept volume approximations of the robot's motion using Swept Spheres Convex Hull, the links of the robot geometry are spherized using FOAM \cite{coumar_foam_2025}. Since the robot occupancy, swept volume, obstacle geometry and its configuration space are modeled as polygonal meshes we leverage MeshLib \cite{noauthor_meshlib_2025} to compute 3D Boolean operations (Intersection, Difference, Union) for partitioning obstacle configuration spaces.

% Precomputing IK solutions for the picks (exclude top grasps due to how it trivializes the problem)

% picture of the 
% Mention adapting APP for continous representation
% 
% Mention Path Seeds

% 

\begin{figure}
    \centering
    \includegraphics[width=0.9\linewidth]{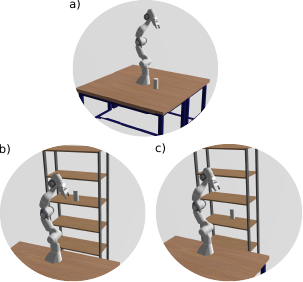}
    \caption{Semi-static environments with 2 movable obstacles, used for evaluating the object-pick tasks. \textbf{(a)} \texttt{Table-Pick} with target object and movable obstacles on a table surface. \textbf{(b)} \texttt{Shelf-High} with target and obstacles on the upper rack. \textbf{(c)} \texttt{ Shelf-Low} with target and obstacles on the lower rack.}
    \vspace{-10pt}
    \label{fig:environments}
\end{figure}

% \todo{adding the blue regions to show the obstacles being on that rack...like, shading the rack with some color on inkscape should be good}

% \todo{Mention that we have fixed start/goals and how many for each experiemnt}

\textbf{Task Environments.} The environments in \autoref{fig:environments}  serve as the primary testbeds for evaluating performance on the object-picking tasks. In \texttt{Table-Pick} (\autoref{fig:environments}(a)), the manipulator, cylindrical object to be grasped, and two movable obstacles are situated on a rectangular table surface. In \texttt{Shelf-High} (\autoref{fig:environments}(b)), the target and obstacles are placed within the upper rack closer to the robot. In \texttt{Shelf-Low} (\autoref{fig:environments}(c)), the target and obstacles are positioned towards the back of the shelf; this introduces narrow passages and depth-reachability constraints that significantly increase planning difficulty.

\subsection{Task Constraints}

\textbf{Goal Specification.} For each environment, we define success by the manipulator's ability to reach any of four candidate pre-grasp poses--\texttt{Pick-Left}, \texttt{Pick-Right}, \texttt{Pick-Front}, \texttt{Pick-Behind}, each denoting the direction from which the grasp target needs to be picked. Top grasps are intentionally excluded to prevent trivializing the grasp task in the \texttt{Table-Pick} environment. We compute and store inverse kinematic (IK) solutions for each pre-grasp pose prior to planning. Due to the geometric constraints of the environments, three pre-grasp poses are valid for \texttt{Table-Pick}, while only two valid pre-grasp poses exist for the \texttt{Shelf-High} and \texttt{Shelf-Low} environments.

\textbf{Movable Obstacles.} Each environment contains two spherical movable obstacles: one fixed at a radius of $0.1$~m, and another that varies across ${0.025, 0.05, 0.075, 0.1}$~m. This yields four obstacle-size pairs per environment: $(0.025, 0.1)$, $(0.05, 0.1)$, $(0.075, 0.1)$, and $(0.1, 0.1)$.

% This is to intentionally render certain goals inaccessible, requiring the planner to identify a collision-free path to any remaining valid candidates.

% We also ensure that, for every movable obstacle, configurations that collide with the target object are removed from its configuration spaces.

% \todo{reviewer says it is not enough if we just exclude the configurations that touch target object, as goals may be invalid...that is exactly what we are going for to make the problem non-trivial and interesting. CC: Explain that in the caption. CC }

\begin{figure*}[!tbp]
    \centering
    \includegraphics[width=\linewidth]{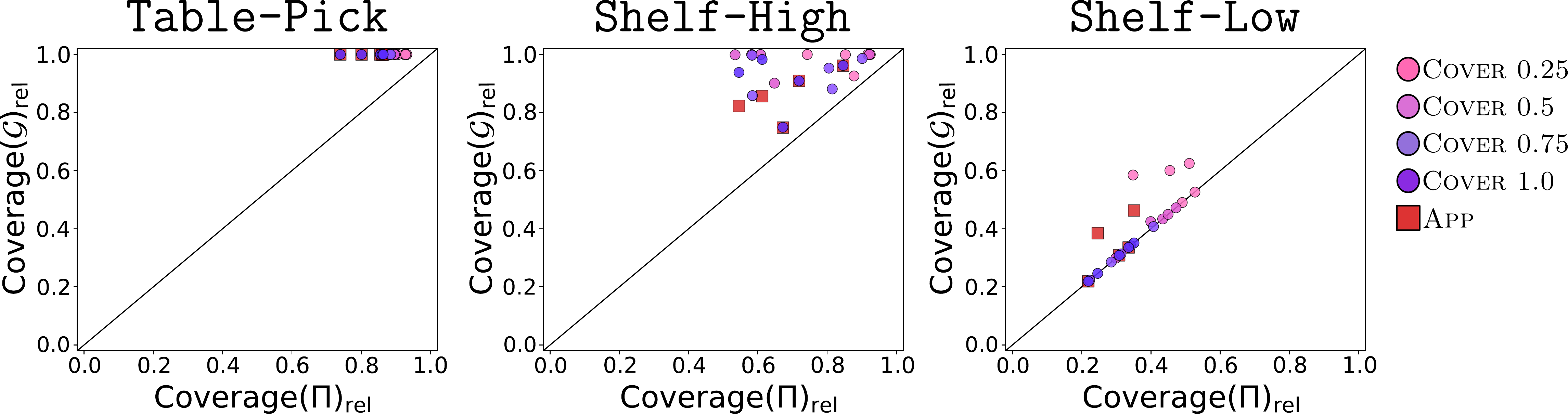}
        \caption{\textbf{Problem-space coverage improvement}. $\text{Coverage}(\graph)_\text{rel}$ achieved by \method vs \app after warm-starting both pipelines with identical seed disjoint paths for multiple trials per experiments. The X-axis denotes the initial relative coverage provided by the seed paths $\text{Coverage}(\Pi)_\text{rel}$, while the Y-axis denotes the relative coverage achieved after the preprocessing phase $\text{Coverage}(\graph)_\text{rel}$. Points above the diagonal line represent coverage improvement over the initial seed paths. Circles indicate individual trial results for \method across varying obstacle-size pairs $o_1, o_2 \in \Om$ (\method 0.25 to 1.0), where the value represents the radius scaling factor of $o_1$ relative to $o_2$. Squares correspond to trial results for \app.}
    \label{fig:results}
    % \vspace{-2em}
\end{figure*}

\subsection{Baseline Comparison}

% To the best of the authors' knowledge, \app and \method are the only existing fixed-time motion planning frameworks designed for semi-static environments. Consequently, \app serves as our primary baseline.

\textbf{Alternative Paths Planner.} We choose \app as our primary baseline, a recent fixed-time motion planning framework designed for semi-static environments. In its preprocessing phase, it computes a set of paths that each remain valid for a subset of arrangements. These are stored in a hash-table, which can be used to retrieve a valid path during the querying phase within constant-time for a given arrangement $m$.

% \app operates by computing $n+1$ disjoint paths\footnote{Two paths are considered disjoint if no configuration of any obstacle can invalidate both paths simultaneously.}, where $n$ is the number of movable obstacles. This is achieved by iteratively solving for composite problems, $p(\mathcal{E}_\Pi)$, where $\mathcal{E}_\Pi \subset \M$ consists solely of arrangements that invalidate previously found paths, $\Pi = \{ \pv_1, \pv_2, \ldots \}$. To distinguish from our formulation of envelopes in \autoref{eqn:envelope}, we refer to $\mathcal{E}_{\pi_i}$ as path-envelope of $\pi_i$, and $\mathcal{E}_\Pi = \bigcup_i\mathcal{E}_{\pi_i}$.

\app operates by computing $n+1$ disjoint paths\footnote{Two paths are considered disjoint if no configuration of any obstacle can invalidate both paths simultaneously.}, where $n$ is the number of movable obstacles. For a path $\pi_i$, \app defines the \emph{path-envelope} $\mathcal{E}_{\pi_i}\subseteq\mathcal{M}$ as the set of arrangements in which all movable obstacles simultaneously invalidate $\pi_i$. Given previously discovered disjoint paths $\Pi=\{\pi_1,\pi_2,\ldots\}$, \app constructs the union envelope $\mathcal{E}_{\Pi}=\bigcup_{\pi_i\in\Pi}\mathcal{E}_{\pi_i}$ and solves the composite problem $p(\mathcal{E}_{\Pi})=\{p(m)\mid m\in\mathcal{E}_{\Pi}\}$, to obtain a path disjoint from the current path set $\Pi$.

% As a result, \app cannot distinguish between the individual free-space contributions of heterogeneous obstacles. In contrast, \method performs arrangement-space partitions for each obstacle separately, making the partition a function of both the robot motion and the individual obstacle's geometry. 

When a sufficient number of disjoint paths cannot be computed, \app recursively subdivides $\mathcal{E}_\Pi$ and solves the resulting composite problems to find alternative overlapping paths. This process continues until the subregions of $\mathcal{E}_\Pi$ are covered by valid paths or a maximum recursion depth is reached. However, the original implementation of \app relies on discretized obstacle configuration spaces for both path-envelope computations and path retrieval from hash-tables.

 % Crucially, this formulation of path-envelopes couples the configuration spaces of all movable obstacles into a single joint representation \M. As a result, \app cannot distinguish between the individual free-space contributions of heterogeneous obstacles. In contrast, \method independently partitions the individual configuration space, $\mathcal{M}_i$, of each obstacle $i$, to capture their individual contributions.

We adapt \app to the continuous domain by redefining its path-envelope computation using the Minkowski sum-based approach employed in \method. Since this change renders the original hash-table retrieval obsolete, we convert the resulting set of paths from \app into a coverage-informed roadmap. This is achieved by repeating \autoref{line:add_path}-\autoref{line:init_filter} in \autoref{alg:cover} to enable valid-path retrieval for the continuous domain. We refer to this adapted version of \app as \appcover.

% Since the original implementation , we adapt \app to continuous spaces by redefining its envelopes via the same geometric backend used in \method (\autoref{alg:decompose}). Both methods are subject to a 10-minute preprocessing timeout and are warm-started with an identical set of disjoint paths to ensure they begin from the same solution base.

\textbf{Planning from Scratch.} We use \rrtconnect as the standard planner baseline. We provide \rrtconnect with a $10$-second timeout, which is double the $5$-second limit given to the \rrtconnect instances used during the preprocessing stages of \method and \appcover.

\subsection{Results and Discussion}
\begin{table*}[h]
\renewcommand{\arraystretch}{1.4}
\setlength{\tabcolsep}{2pt}
\centering
% \caption{Performance Comparison: Problem Space Coverage () and Query Time, SR achieved with \method vs \app vs \pfs}
\caption{\textbf{Performance comparison.} Problem-space coverage and query benchmarking results for \method, \app, and \pfs across environments with varying movable obstacle sizes. 
$t_\text{query}$ denotes query time in milliseconds. 
$|V|$, $|E|$, and $|\pv|$ denote the number of vertices, edges, and paths in \graph. 
$\mathrm{SR}_\text{query}$ is the success rate over $1000$ motion-planning queries. 
$\text{Coverage}(\graph)_\text{rel}$ denotes relative problem-space coverage, defined as $\text{Coverage}(\graph) / \text{Coverage}(\graph)_\text{max}$, where $\text{Coverage}(\graph)_\text{max}$ excludes obstacle arrangements in $\Om$ that invalidate all start or goal configurations.}

\label{table:query_results}
\small
\begin{tabular*}{\textwidth}{@{\extracolsep{\fill}}l cccc cccc cc @{}}
\toprule
\multirow{2}{*}{\textbf{Scenario ID}} & \multicolumn{4}{c}{\textbf{\method}} & \multicolumn{4}{c}{\textbf{\appcover}} & \multicolumn{2}{c}{\textbf{\pfs}} \\
\cmidrule(lr){2-5} \cmidrule(lr){6-9} \cmidrule(lr){10-11}
 & $t_\text{query} (\text{ms}) $ & $|V|, |E|, |\pi|$ & $\mathrm{SR}_\text{query} $ & $\text{Coverage(\graph)}_\text{rel}$ & $t_\text{query} (\text{ms}) $ & $|V|, |E|, |\pi|$ & $\mathrm{SR}_\text{query}$ & $\text{Coverage(\graph)}_\text{rel}$ & $t_\text{query} (\text{ms})$ & $\mathrm{SR}_\text{query}$ \\
\midrule
Table-0.25 & $\mathbf{10.86 \pm 4.69}$ & 14, 16, 4 & \textbf{100 \%} & \textbf{0.995} & $64.79 \pm 19.16$ & 43, 51, 10 & 99.5 \%  & 0.989 & $1408 \pm 445$ & \textbf{100 \%} \\
Table-0.5  & $\mathbf{22.20 \pm 9.38}$ & 17, 20, 8 & \textbf{100 \%}  & \textbf{0.994} & $64.79 \pm 19.16$ & 43, 51, 10 & 99.5 \%  & 0.989 & $1347 \pm 431$ & \textbf{100 \%} \\
Table-0.75 & $\mathbf{7.34 \pm 1.35}$ & 12, 14, 5 & \textbf{100 \%}  & \textbf{0.993} & $64.79 \pm 19.16$ & 43, 51, 10 & 99.5 \%  & 0.989 & $1260 \pm 445$ & 99.9 \%  \\
Table-1.0  & $\mathbf{8.80 \pm 1.38}$ & 12, 15, 7 & \textbf{100 \%}  & 0.988 & $64.79 \pm 19.16$ & 43, 51, 10 & 99.5 \%  & \textbf{0.989} & $1306 \pm 473$ & 99.9 \% \\
\midrule
ShelfH-0.25 & $10.62 \pm 4.60$ & 26, 26, 3 & \textbf{100 \%}  & \textbf{0.967} & $\mathbf{9.41 \pm 4.44}$ & 18, 18, 2 & \textbf{95.9 \%} & 0.883 & $7115 \pm 2410$ & 36.4 \%  \\
ShelfH-0.5  & $18.17 \pm 7.80$ & 24, 27, 26 & \textbf{99.5 \%}  & \textbf{0.939} & $\mathbf{9.41 \pm 4.44}$ & 18, 18, 2 & \textbf{95.9 \%} & 0.883 & $7229 \pm 2583$ & 38.1 \% \\
ShelfH-0.75 & $16.56 \pm 5.97$ & 20, 20, 4 & \textbf{98.1 \%}  & \textbf{0.919} & $\mathbf{9.41 \pm 4.44}$ & 18, 18, 2 & \textbf{95.9 \%} & 0.883 & $7186 \pm 2549$ & 35.4 \%  \\
ShelfH-1.0  & $10.58 \pm 4.81$ & 17, 20, 5 & 87.3 \%  & \textbf{0.890} & $\mathbf{9.41 \pm 4.44}$ & 18, 18, 2 & \textbf{95.9 \%} & 0.883 & $7038 \pm 2733$ & 31.9 \%  \\
\midrule
ShelfL-0.25 & $\mathbf{7.98 \pm 1.37}$ & 21, 22, 3 & \textbf{67.2 \%}  & \textbf{0.617} & $18.7 \pm 6.2$ & 18, 18, 2 & 50.3 \%  & 0.432 & $7589 \pm 1842$ & 10.1 \%  \\
ShelfL-0.5  & $\mathbf{7.15 \pm 3.42}$ & 15, 14, 1 & \textbf{50.9 \%}  & \textbf{0.454} & $18.7 \pm 6.2$ & 18, 18, 2 & 50.3 \%  & 0.432 & $7523 \pm 2033$ & 8.7 \%  \\
ShelfL-0.75 & $\mathbf{8.33 \pm 4.02}$ & 12, 11, 1 & 43.4 \%  & 0.384 & $18.7 \pm 6.2$ & 18, 18, 2 & \textbf{50.3 \%}  & \textbf{0.432} & $7891 \pm 1491$ & 7.7 \%  \\
ShelfL-1.0  & $\mathbf{15.8 \pm 4.6}$ & 13, 12, 1 & 38.5 \%  & 0.324 & $18.7 \pm 6.2$ & 18, 18, 2 & \textbf{50.3 \%}  & \textbf{0.432} & $7749 \pm 2079$ & 6.0 \%  \\
\bottomrule
\end{tabular*}
\end{table*}

\textbf{Problem-space Coverage.} We evaluate the problem-space coverage achieved by \method and \appcover under a fixed preprocessing budget of 10 minutes. Each experiment is repeated across five independent trials, and the maximum coverage achieved across trials is reported. To ensure a fair comparison, both pipelines are warm-started using the same set of initial disjoint paths $\Pi$ as shown in \autoref{fig:results}. Coverage of the computed roadmaps is reported using the relative coverage metric $\mathrm{Coverage}(\graph)_\mathrm{rel}$, which measures the fraction of the feasible arrangement space that admits a valid path in $\graph$.

The path-envelope formulation used in \app does not distinguish how different obstacles contribute to each $\mathcal{E}_{\pi_i}$. Consequently, for heterogeneous obstacle pairs, the envelope is effectively governed by the largest obstacle, reducing such cases to the homogeneous case defined by the largest radius. Under \appcover, all heterogeneous pairs therefore collapse to the case $(0.1\,\mathrm{m}, 0.1\,\mathrm{m})$, and the same results are reported across those scenarios. In contrast, \method maintains independent partitions for each obstacle configuration space and can therefore evaluate heterogeneous obstacle pairs distinctly.

\textbf{Querying.} To evaluate query-time performance, we sample $1000$ obstacle arrangements per environment and measure the retrieval time $t_{\text{query}}$ and the success rate $\mathrm{SR}_{\mathrm{query}}$ of each method. Arrangements that invalidate all start configurations or all goal configurations are rejected, since such instances are infeasible regardless. The success rate $\mathrm{SR}_{\text{query}}$ is defined as the fraction of sampled queries for which the planner returns a valid path.

% \todo{do we alos talk about abysmal planning times for eveyrthing because we are hampered by the extra collsiion checking with meshlib?}

\textbf{Table-Pick.} \autoref{table:query_results} shows that both \method and \appcover achieve near-complete coverage\footnote{To maintain mutually exclusive obstacle-space partitions, meshes are slightly eroded after each partition during BSP tree construction, resulting in a small cumulative loss of measurable volume and causing the reported coverage to fall marginally below 1.0.} in this environment. The \texttt{Table-Pick} is the least constrained environment, where the robot operates on an open tabletop surface with relatively few geometric restrictions around the target object. This is corroborated by the low planning times and near-perfect success rate achieved by \rrtconnect when planning from scratch. \method consistently achieves $\mathrm{Coverage}(\graph)_{\mathrm{rel}} > 0.98$ across all obstacle pairs, while \appcover achieves comparable coverage. \autoref{fig:results} shows limited improvement beyond the initial disjoint paths, indicating that they already capture most feasible arrangements.

\textbf{Shelf-High.} The presence of the shelf introduces tighter geometric constraints around the target object. \rrtconnect succeeds in only $32$--$38\%$ of queries, indicating the increased difficulty of this setting. Despite this increase in problem difficulty, both \method and \appcover maintain high $\mathrm{SR}_{\mathrm{query}}$ during querying. Under these conditions, the disparity between \method and \appcover becomes more apparent for heterogeneous obstacle pairs. 

\textbf{Shelf-Low.} Here the target object lies deeper within the shelf, making this the most restrictive setting. \rrtconnect succeeds only $6$--$10\%$ of queries. Both \method and \appcover achieve moderate query success ($\approx 55$--$75\%$). \autoref{fig:results} further shows that neither method expands coverage beyond the initial disjoint seed paths in this setting. \appcover achieves marginal improvements in a few trials due to its recursive subdivision of path-envelopes when composite problems fail, while \method does not expand beyond the seed paths. However, \method provides larger coverage for heterogeneous cases involving smaller obstacles.

Empirical query success rates closely follow the achieved roadmap coverage in each experiment. Across all environments, both \method and \appcover achieve millisecond-scale query times ($t_{\mathrm{query}}$). Within an environment, $t_\mathrm{query}$ grows with the roadmap size ($|E|, |\Pi|$). These results demonstrate the advantages of \method. Once coverage is verified, queries can be answered in milliseconds with fixed-time guarantees, whereas planning from scratch with \rrtconnect requires seconds per query.

\section{Discussion}
We presented \method, a roadmap construction method that reasons about
problem-space coverage in semi-static environments. By associating roadmap edges with obstacle-space partitions, \method explicitly certifies feasible subsets of the arrangement space, moving beyond heuristic or purely discretization-based approaches. This representation enables coverage across heterogeneous obstacle sizes and placements while retaining fixed-time query guarantees.

Several limitations remain. Preprocessing cost grows with the number of
obstacles and roadmap edges, which may limit scalability in larger
environments. In addition, the current implementation does not recursively refine unresolved composite problems, and therefore cannot certify that uncovered partitions are infeasible, except in degenerate cases where all starts or all goals are invalid. Incorporating infeasibility-detection methods \cite{li_sampling_2023} and principled recursive refinement strategies is a promising direction. More broadly, \method may serve both as a diagnostic tool for evaluating roadmap coverage and as a foundation for more efficient Task and Motion Planning frameworks \cite{garrett2021integrated}.

\bibliographystyle{IEEEtran}
\bibliography{refs_no_url}

\end{document}